# Behaviorally Grounded Model-Based and Model Free Cost Reduction in a Simulated Multi-Echelon Supply Chain

James Paine[*]

Massachusetts Institute of Technology Sloan School of Management

December 18, 2021

**Abstract:** Amplification and phase shift in ordering signals, commonly referred to as 'bullwhip', is responsible for both excessive strain on real world inventory management systems, stock outs, and unnecessary capital reservation though safety stock building. Bullwhip is a classic, yet persisting, problem with reverberating consequences in inventory management. Research on bullwhip has consistently emphasized behavioral influences for this phenomenon and leveraged behavioral ordering models to suggest interventions. However more recent model-free approaches have also seen success. In this work, the author develops algorithmic approaches towards mitigating bullwhip using both behaviorally grounded model-based approaches alongside a model-free dual deep Q-network reinforcement learning approach. In addition to exploring the utility of this specific model-free architecture to multi-echelon supply chains with imperfect information sharing and information delays, the author directly compares the performance of these model-based and model-free approaches. In doing so, this work highlights both the insights gained from exploring model-based approaches in the context of prior behavioral operations management literature and emphasizes the complementary nature of model-based and model-free approaches in approaching behaviorally grounded supply chain management problems.

---

[*] E62-441, 100 Main Street, Cambridge, MA 02142. Email: jpaine@mit.edu

# 1 Introduction and Background

The field of operations management has increasingly followed its peers in economics, marketing, and finance by endeavoring to recognize the influence of human heuristic-based decision rules and incorporate these behavioral observations into the models of supply chains and inventory management (Gino & Pisano, 2008). Among one of the more studied consequences of the interaction of behavioral heuristics and supply chain structure is the emergence of instability as embodied by the 'bullwhip effect' (Croson et al., 2014; Forrester, 1961; Lee et al., 2004). Bullwhip refers to the increasing amplitudes in both orders and on-hand inventory positions of members of a multi-echelon supply chain the further one moves away from a source of order variability.

The bullwhip effect is responsible for both excessive strain on real world inventory management systems, stock outs, and unnecessary capital reservation though safety stock building (Ellram, 2010). This phenomena is also not necessarily restricted to any one industry, but rather present in varying forms whenever ordering decisions being made in moderately complex and interlinking environments such as multi-echelon supply chains (Lee et al., 2004; Sterman, 1989b, 2000). And while often viewed as a 'classical' problem in Operations Management, recent worldwide experiences with supply chain shortages and excesses induced by the Coronavirus pandemic for consumer goods, foodstuffs and medical supplies, have catapulted the term 'bullwhip' into the popular consciousness (see for example: Bamakan et al., 2021; Evenett, 2021; Hockett, 2020; Johnson, 2021; Shih, 2020; Stank et al., 2021)

Optimal control policies in multi-echelon supply chains are well understood and well-studied. Work by Clark and Scarf demonstrated that an optimal control policy can be applied via a base-stock ordering system when the final customer demand distribution is known (Clark & Scarf, 1960). Their algorithm was later generalized and operationalized to both multi-echelon supply chains with imperfect local information and stationary demand patterns (Chen, 1999; Chen & Samroengraja, 2009; Lee et al., 2004).

Behavioral causes of ordering and inventory amplification have also been thoroughly explored. A key behavioral bias that leads to bullwhip is commonly identified as 'supply-chain underweighting' (Croson & Donohue, 2006; Narayanan & Moritz, 2015; Sterman, 1989a) and emerges as part of a larger anchoring and adjustment heuristic employed by decision makers in an multi-echelon supply chain (Sterman, 1989a; Tversky & Kahneman, 1974). Mitigation of bullwhip has focused on adjusting both the structure of the supply chain itself and the information availability along the supply chain (Croson et al., 2014; Croson & Donohue, 2006; Wu & Katok, 2006), and on the instruction and training strategies of supply chain managers (Croson et al., 2014; Martin et al., 2004; Wu & Katok, 2006). While mitigation is possible, the underlying ordering heuristics that drive the emergence of bullwhip remain in many of these studies.

These prior explorations of the causes of ordering amplification and phase shift, and the resulting suggested interventions, rely on developing an underlying model of human ordering behavior (as in the case of the Sterman, Corson & Donohue, and Wu and Katok work) or on explicitly modeling the supply chain structure and information network (as in the case of the classical Clark and Scarf or even the more recent Hau work). Recently years have seen increased success by abandoning these model-based approaches and embracing a model-free reinforcement learning approach to minimizing the costs associated with bullwhip in a model of a multi-echelon supply chain. These model free approaches either modify the more traditional models of a multi-echelon supply chain to allow full information sharing (Chaharsooghi et al., 2008; Thompson & Badizadegan, 2015), or more recently employ a modified Deep Q-Network (DQN) approach to a setting with more limited information sharing (Oroojlooyjadid et al., 2021).

The emergence of these two approaches, both model-based and model-free, presents an opportunity to contrast and compare these two fundamentally different frameworks to bullwhip mitigation. Additionally, by considering these two approaches together, it allows for a more direct examination of the resulting ordering policies in the context of exiting behavioral literature, even in the model-free contexts.

Most promisingly, this work opens an avenue to create a *useful*, *implementable*, and *understandable* policy interventions capable of *mitigating* bullwhip generated by real humans when placed into an actively evolving inventory management crisis in-progress. In this manner this work strives to 'close the loop' first begun by those endeavoring to define behavioral operations management. In this prior work, the authors define 'Behavioral Operations Management', in part, by incorporating observations of human decision heuristics into operations management optimization strategies (Gino & Pisano, 2008; Größler et al., 2008).

In this paper, by developing methods to minimize cost along a supply chain and mitigate the bullwhip effect, that are *also* directly interpretable in the context of existing models of human decision making, it is possible to map from optimization routines that minimize cost functions which incorporate the human environment back to interpretable human-modeled decision rules. In doing so this work suggests features of operations systems, specifically at the information and physical flow interfaces of entities with those systems, that help minimize distortions and costs.

## 2 Methods and Model Development

For many of the above referenced studies, the Beer Game (Sterman, 1989a) is the modeling framework employed to explore and test the interventions developed. The Beer Game provides an ideal, and well-studied environment for use in this work as well. The Beer Game is a classical inventory management and System Dynamics simulation and learning tool, in which a multi-agent decentralized supply chain is modeled, much like real decentralized inventory management systems). First developed by Jay Forrester at MIT, the game has been used since the 1950's to illustrate system thinking concepts and the prevalence of the bullwhip effect. The original purpose of the simulation was to illustrate the difficulty of rational thinking in the midst of time-delayed and non-linear information feedback loops, value of information sharing, and most classically the bullwhip effect in inventory management (Sterman, 1989a, 1989b).

Figure 2 below shows the typical starting layout for the game as used in numerous previous studies using this modeling framework (Croson & Donohue, 2006; Narayanan & Moritz, 2015; Sterman, 1989a), which is started with 12 units of inventory on hand for each player, and 4 units of inventory in transit at each stage in the shipping system, and 4 orders moving through the order chain. The original purpose of the simulation was to illustrate the difficulty of rational thinking in the midst of time-delayed and non-linear information feedback loops, value of information sharing, and most classically the bullwhip effect in inventory management (Sterman, 1989a, 1989b).

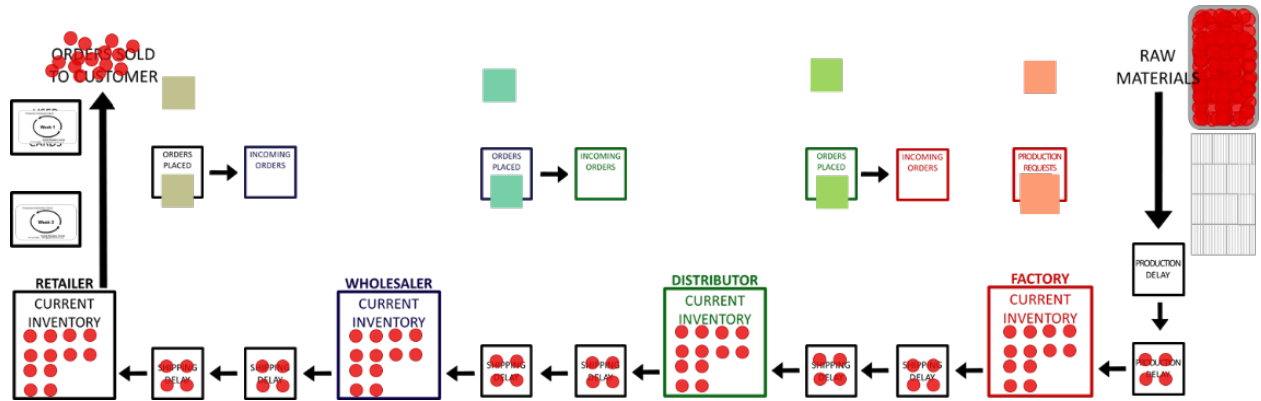

**Figure 1. Example of Beer Game Board Layout**

Following the same examples from pervious uses of the Beer Game as a model of a multi-echelon supply chain, after four training rounds (where in the customer demand is a fixed quantity of 4 units, and all players are directed to place an order of 4 units downstream), the customer orders experience a stepwise increase (from 4 units to 8 units per round). From this point onwards, each round of the game proceeds as follows:

1. Receiving inventory and advance shipping delays – Each entity receives the units in the shipping delay immediately to their right. The contents of the furthest shipping delay to the right are moved up
2. Fill orders – Entity 1 (retailer) views the customer order, all others examine the 'incoming orders' and orders, inclusive of any outstanding backorders, are filled to the extended inventory allows
3. Record inventory or backlog
4. Advance order slips – the order slips further to the left are moved up
5. Place orders – Each entity decides what to order and places it the 'orders placed' box to their right

The stated goal of the game is to reduce the amount of *total cost of the entire team* over some time horizon *T,* subject to some known inventory holding and backorder/stockout costs. Backorders do not expire under the traditional interpretation of this game and must be filled from existing stock prior to meeting any new demand. In the prior studies referenced above, and in this work, the cost of holding inventory, $C_{inv}$, is $0.50 per unit per period, and the cost of backorders, $C_{bo}$, is $1.00 per unit per period.

$$Cost_{Team} = \sum_{t=1}^{T} \sum_{entity=1}^{N} (C_{bo} * Backorders_{t,n} + C_{inv} * Inventory_{t,n}) \qquad (1)$$

Typically, real human players are placed into this system to make inventory purchasing and management decisions. Within a few rounds of ordering, the bullwhip in inventory and backorders appears, amplifying over time along the simulated supply chain as each player acts to reserve inventory to satisfy their own myopic forecasts and needs. Exact solutions for optimal ordering quantities have been developed, such as the base-sock method (Chen & Samroengraja, 2009; Clark & Scarf, 1960), but require all agents to be acting rationally and consistently. Additionally, while these optimal ordering methods presume stationary customer order patterns, which this simulation satisfies, the human participants themselves have no knowledge *a priori* of the distribution of the customer order pattern.

A discrete time model of the Beer Game, following the sequence of steps described in (Sterman, 1989a), was created in the R and Pythons scripting languages. This model was made as both a self-contained simulation of the system over a given time horizon, and as a callable function that takes a given state-action pair and returns an updated state, given an ordering rule for the entities in the system. This ordering rule has been the subject of much research in the past, but for this work is taken to be the ordering heuristic introduced in (Sterman, 1989a) and (Martin et al., 2004), and summarized in expression (2) and (3) below:

$$O_t = MAX(0, \hat{L}_t + \alpha_S(S' - S_t - \beta\, SL_t) + \varepsilon_t) \qquad (2)$$

$$\text{where } \hat{L}_t = \theta L_t + (1-\theta)\hat{L}_{t-1} \qquad (3)$$

In the above, $O$ is the order placed at time $t$ given the information observed in the right-hand side of the above expression. In that expression $\hat{L}$ is a smoothed interpolation of the expected outflow of inventory, subject to a smoothing parameter $\theta$. *SL* refers to the total inbound supply line of inventory heading towards the player. *S* is the current on-hand inventory (or stock), and *S'* is a parameter that can be considered analogous to the desired or goal on-hand inventory of the player. Thus, we have an expression with four parameters: θ, α, β, and S'. As conceptualized in (Sterman, 1989a), the above parameters are bounded as $0 \leq \theta, \alpha, \beta \leq 1$ and $0 \leq S'$, and that paper also provides fitted values for expression (2) and (3) for a set of real human teams, along with a set of parameter values that best fit the overall behavior of all teams. This model of a multi-echelon supply chain and the corresponding estimated behavior of humans in that system serves as the testing bed for subsequent agent development and optimizations.

## 2.1 Reducing Costs Via a Model of Human Ordering Heuristics

Expressions (2) and (3) are ultimately a model of human behavior in a multi-echelon supply chain and this project first focuses on the results of minimizing the costs incurred by the entire team by fixing the ordering parameters of all entities save one in a nested set of these expressions, and finding a set of parameters for the remaining entity that are cost reducing. This routine is illustrated in Figure 2.

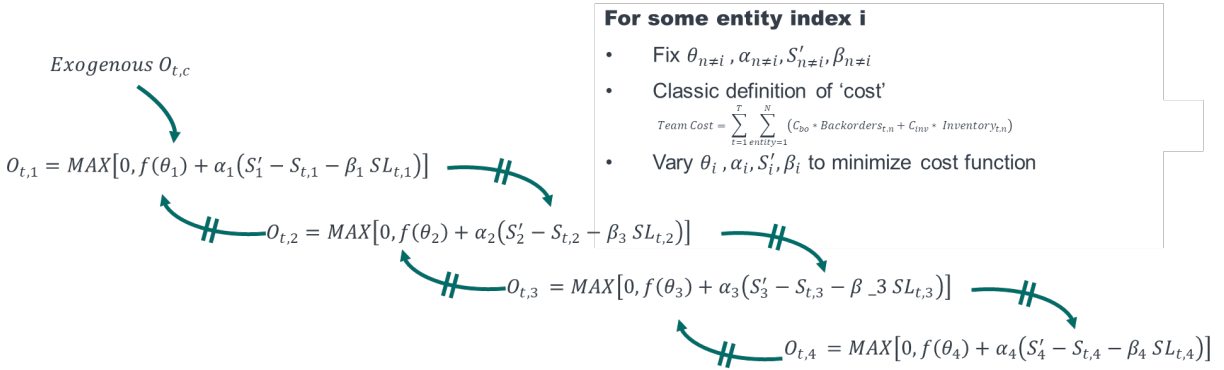

**Figure 2. Cost Minimization Routine for the Model-Based Approach**

To test robustness, the parameters for the other simulated players were drawn from a variety of combinations of available values, however the general results are the similar independent of the choice of parameters from the feasible (and previously fitted from real data) set. An illustrative example of the reduction in ordering amplification for an agent acting in the 'Distributor' position of the supply chain is shown below in Figure 3 relative to the 'general' case set of parameters.

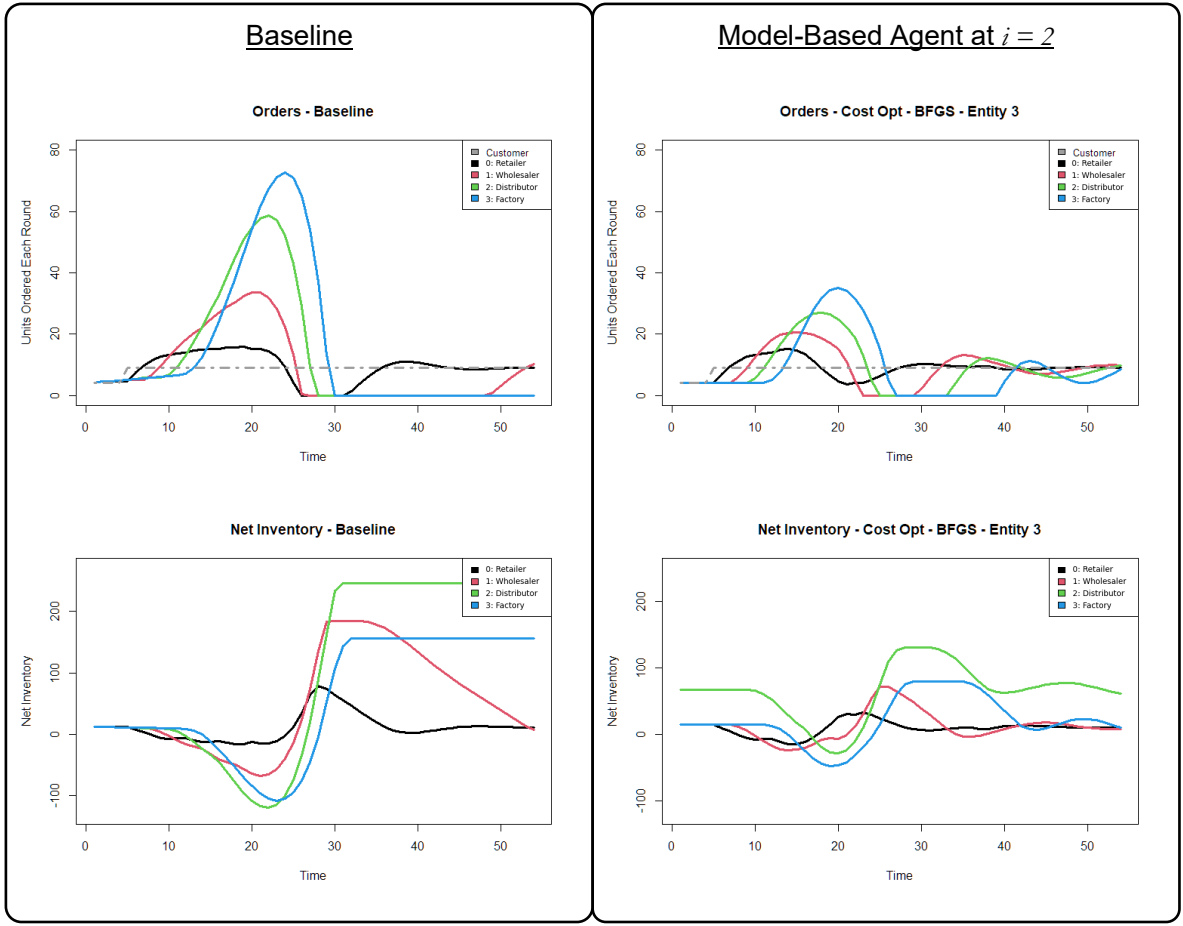

**Figure 3. Model-Based Cost Reduction: Baseline Costs = 9978, Reduced Costs = 3225 (-67%)**

Furthermore, the general learned parameters for a horizon of $t = 52$ periods utilizing a bounded BFGS method to reduce costs (Byrd et al., 2005) is shown in Table 1.

**Table 1. Learned Parameters for the Model-Based Cost Reducing Agent**

| Entity Optimized | Fitted Parameters | | | | Total Cost (Inventory-Based) |
|---|---|---|---|---|---|
| | θ | α | β | S' | |
| N/A (baseline) | 0.3600 | 0.2600 | 0.3400 | 17.0000 | 9978.44 |
| 0 (Retailer) | 0.002 | 0.409 | 0.975 | 29.259 | 1440.45 (-85.56%) |
| 1 (Warehouse) | 1.000 | 0.495 | 1.000 | 36.405 | 1911.77 (-80.84%) |
| 2 (Distributor) | 0.747 | 0.094 | 0.784 | 73.721 | 3225.41 (-67.68%) |
| 3 (Factory) | 1.000 | 1.000 | 0.048 | 21.581 | 4799.45 (-51.9%) |

## 2.2 Reducing Costs Via a Model-Free DQN Approach

The model-free approach utilizes reinforcement learning, which at its core features a 'perception-action-learning' loop (Sutton & Barto, 2014) in which a model-free agent interacts repeatedly with its environment in order to construct an action policy that maximizes received rewards. The model-based methods above presume that the agent has some knowledge of the environment dynamics and can leverage this knowledge to make predictions of action outcomes. Said more directly, the model-based agent developed above has some a priori knowledge of the transition probabilities and the resultant expect rewards from the underlying Markov decision process. The model-free approach in contrast has no pre-existing idea of relationship between actions, states, and rewards. This agent does not estimate transition probabilities but rather estimates reward and observation of the current state. Additionally, as this method had no pre-exiting assumptions of the underlying dynamics of the system, it may possibly be more cross-applicable to other complex, but similar, environments or one with more stochastic elements.

The Beer Game itself has been used as a training environment in reinforcement learning applications, most notably utilizing various modifications of Deep Q-Network architectures, including modifications to allow for independent training across entities utilizing pooled reward schemes (Chaharsooghi et al., 2008; Opex Analytics, 2018; Oroojlooyjadid et al., 2021). The Beer Game as a model of a multi-echelon supply chain presents a challenge to direct application of DQN architecture, challenges that are often also found in real integrated supply chains. Specifically challenges emerge from 1) the true 'full state' of information is unknown to any one entity, 2) rewards are communal and not realized until the end of the time horizon, 3) DQN architectures can be 'over optimistic' in even mildly noisy environments (Thrun & Schwartz, 1993), and perhaps most importantly 4) the current overarching quality of the system, e.g. whether bullwhip is in progress or if the supply chain is stable, matters almost as much if not more than any specific action, and

In order to address the above issues, most notably the final point in the above list, this work presents a DQN architecture for use in multi-echelon supply chains like the Beer

Game that has the following general architecture: 1) An 'order-plus' action space (Oroojlooyjadid et al., 2021) which both allows for unbounded ordering in absolute terms and follows from observations in the model-based approach above, 2) a dual DQN network (Wang et al., 2016) that separately maintains a value function estimation for both the current overarching combined state of the system and separately for each action, 3) an observation space defined over a window of prior state observations corresponding to the signal delay in the system, 4) a combination of epsilon-greedy and Boltzmann exploration policies (Wiering, 1999), and finally 5) three sequential dense layers with ReLu activations.

      The environment itself is built on the same framework utilized in the model-based approach above, with the functionalized form of the Beer Game translated into the commonly used opensource Gym research framework developed by OpenAI (Brockman et al., 2016)[1]. This environment allows for training against all positions in the simulated supply chain against randomly bootstrapped assemblages of human-like players, whose ordering rules are drawn from classical supply chain literature (Sterman, 1989a). Additionally, this agent can be trained against random (but bounded) simulation horizons to avoid over-learning endgame-dependent policies, and even noisy realizations of ordering decisions. An illustration of this framework is shown in Figure 4.

---

[1] Supporting files, including the code used to generate the results in this manuscript can be found online at https://github.com/jpain3/Taming-the-Bull

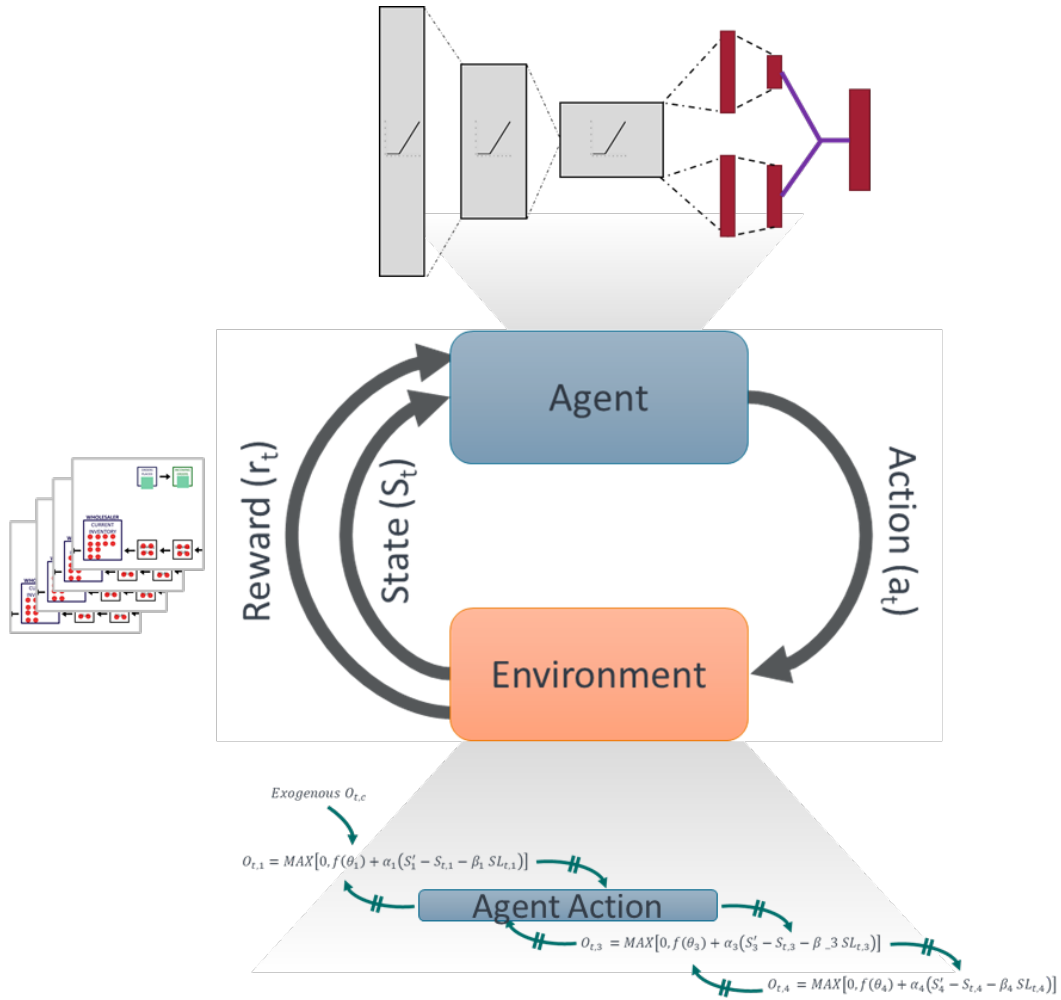

**Figure 4. Cost Minimization Framework for the Model-Free DQN Approach**

As with the model-based approach, the DQN network developed above is generally able to develop a policy that can reduce costs along the supply chain. A single example of this, for an agent acting in the 'wholesaler position' of the supply chain relative to the general set of parameters is shown in Figure 5.

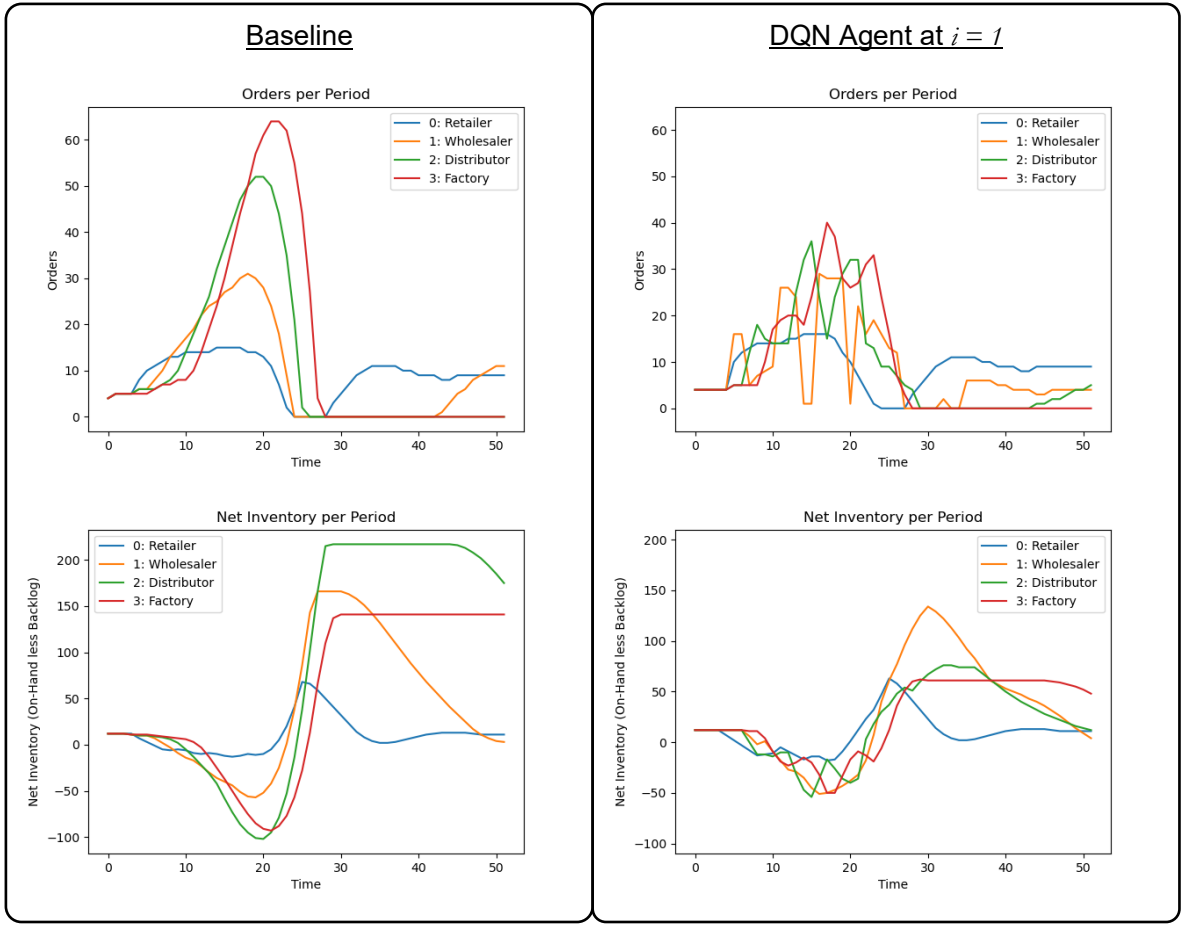

**Figure 5.** Model-Free Cost Reduction: Baseline Costs = 8547, Reduced Costs = 3765 (-56%)

## 2.3 Comparing Model-Based and Model-Free Approaches

The development of both approaches above using the same underlying modeling framework allows for a probing of the boundary between model-based and model-free approaches. The section above presents results that shown that both methods can be cost reducing, but the true value of this work lies in the comparison of these methods more so than the separate observation of their overarching utilities. The resultant trained agents, both model-based and model-free, were exposed to scenarios that are increasingly outside of their original training environments by first randomly drawing team members based on 'real' human players from prior work and injecting increasingly noisy realization of ordering decisions by these other players. For these preliminary results, the randomly assembled fellow 'players' are drawn from ordering models for 11 'real' team members from Sterman '89. For each period, adjust ordering decision for other players by $O_{t,i} = \hat{O}_{t,i} + N(0, \sigma^2)$, and then perform 500 repetitions both with and without (as a baseline average) the trained agent in play. This was repeated for each entity position in the supply chain while varying $\sigma \in [0,15]$.

# 3 Preliminary Results

Table 2 shows selected results of the average cost reduction achieved by each agent type at each position in the supply chain under increasingly noisy realizations of the ordering heuristics used by the other simulated actors in the system.

**Table 2. Cost Reduction versus Baseline on Average over 500 Runs with Increasing Order Noise**

| | | Model-Based Approach | | | | Model-Free DQN Approach | | | |
|---|---|---|---|---|---|---|---|---|---|
| | | Supply Chain Position | | | | Supply Chain Position | | | |
| | | 0 (Retailer) | 1 (Wholesaler) | 2 (Distributor) | 3 (Factory) | Retailer | Wholesaler | Distributor | Factory |
| $\sigma$ as used in $O_{t,i} = \hat{O}_{t,i} + N(0, \sigma^2)$ | 0.0 | -46% | -25% | -57% | -35% | -34% | -46% | -40% | 30% |
| | 0.5 | -43% | -32% | -58% | -36% | -31% | -51% | -43% | 26% |
| | 1.0 | -40% | -29% | -57% | -35% | -31% | -53% | -42% | 25% |
| | 1.5 | -40% | -24% | -55% | -29% | -37% | -54% | -44% | 30% |
| | 2.0 | -35% | -24% | -55% | -27% | -28% | -57% | -46% | 28% |
| | 2.5 | -26% | -19% | -55% | -26% | -27% | -57% | -47% | 27% |
| | 3.0 | -21% | -19% | -53% | -28% | -24% | -58% | -49% | 18% |
| | 3.5 | -25% | -19% | -53% | -25% | -28% | -58% | -51% | 15% |
| | 4.0 | -13% | -15% | -46% | -22% | -17% | -60% | -46% | 13% |
| | 4.5 | -7% | -5% | -44% | -21% | -11% | -55% | -46% | 15% |
| | 5.0 | -14% | -6% | -46% | -20% | -21% | -56% | -49% | 8% |
| | 5.5 | -11% | -3% | -43% | -17% | -19% | -51% | -48% | 7% |
| | 6.0 | -2% | 0% | -38% | -16% | -11% | -49% | -45% | 7% |
| | 6.5 | -7% | 6% | -36% | -19% | -7% | -47% | -42% | 1% |
| | 7.0 | -11% | 5% | -31% | -9% | -15% | -46% | -39% | 5% |
| | 7.5 | -6% | 13% | -34% | -6% | -6% | -40% | -42% | 5% |
| | 8.0 | 0% | 11% | -30% | -7% | -2% | -40% | -40% | 4% |
| | 8.5 | 2% | 13% | -33% | -7% | -1% | -38% | -43% | -1% |
| | 9.0 | -6% | 23% | -23% | -2% | -3% | -30% | -35% | 1% |
| | 9.5 | 0% | 14% | -23% | -3% | 0% | -35% | -35% | 1% |
| | 10.0 | -6% | 36% | -25% | -2% | -1% | -26% | -37% | 1% |
| | 10.5 | 0% | 27% | -21% | -7% | 0% | -30% | -33% | -5% |
| | 11.0 | 7% | 32% | -22% | -1% | 9% | -25% | -34% | -1% |
| | 11.5 | -2% | 41% | -23% | -5% | 3% | -22% | -36% | -8% |
| | 12.0 | -3% | 31% | -17% | 3% | 2% | -21% | -31% | 0% |
| | 12.5 | 7% | 25% | -18% | -2% | 9% | -23% | -31% | -4% |
| | 13.0 | 0% | 34% | -14% | 2% | 7% | -22% | -28% | -5% |
| | 13.5 | 1% | 41% | -14% | 0% | 9% | -19% | -28% | -4% |
| | 14.0 | 1% | 43% | -15% | 4% | 11% | -15% | -29% | -3% |
| | 14.5 | 8% | 28% | -9% | -3% | 13% | -21% | -25% | -11% |
| | 15.0 | -4% | 40% | -9% | 5% | 4% | -16% | -24% | -4% |

While the above clearly shows deceased performance across both agent types under increasing stochastic conditions, it also helps illustrate the positions in the supply chain in which each approach may have a relative advantage. To make this observation clearer, Figure 6 shows the relative advantage the model-free DQN agents have versus the model-based agents in reducing costs at different positions in the supply chain and increasingly noisy realizations of the heuristic ordering rules.

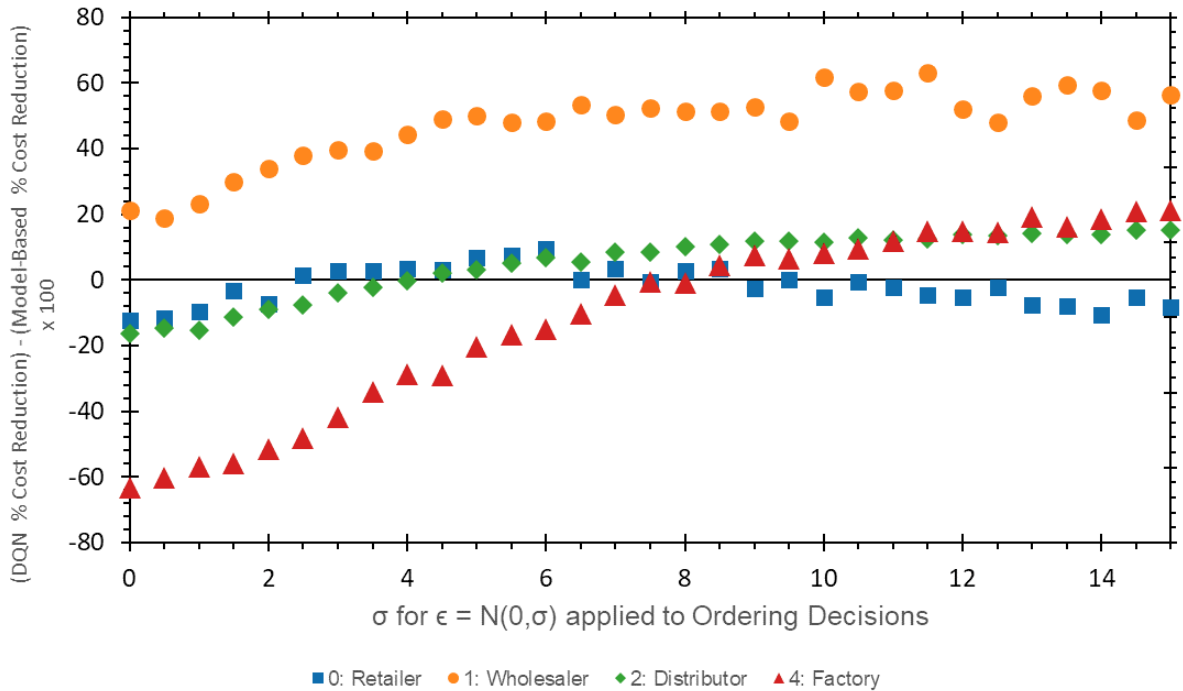

**Figure 6. Relative Cost Reduction Advantage of DQN Model-Free Agent vs Model-Based Agent**

# 4 Discussion

As mentioned in the Introduction, one goal of this project is to develop, wherever possible, *interpretable* agents. The resulting parameters from the model-based agents shown in Table 1 allow for interpretation in the context of prior behavioral operations management literature and provides interesting support for that prior work. Notably:

## 4.1 Low values of θ for the Retailer and high values of θ for everyone else:

This parameter determines the degree of smoothing in updating each entity's expectation of future orders in the same manner as the classic anchoring and adjustment heuristic (Tversky & Kahneman, 1974). For low values of θ, the entity is slow to update expectations while for high values of θ, the entity is quick to adopt the new order signal being received as their expectation for the future. Here, low values of θ for the Retailer, or Entity 1, means that this entity which is most downstream in the chain and most influential towards information flow upstream to other supply chain partners, is slow to update their expectation of changes in customer orders and thus unlikely to rapidly change order signals. Conversely, the high (often at or near 1) values of θ for downstream entities can be interpreted as a high level of trust in the order signals being sent from upstream partners. As discussed in prior research, trust is an essential part of a well-functioning supply chain and some degree of trustworthiness must be assumed in a well function integrated supply chain (Özer et al., 2011). The values of θ found here imply that bullwhip minimization is achieved, in part, by cautious response to changes in order signals from customers, but full trust in order signals from partners.

## 4.2 Very high values of β at all positions in the supply chain:

All entities optimized to minimize the cost of the system in the presence of simulated human-like partners did so in part by *not* falling prey to the supply chain underweighting heuristic observed in previous empirically-grounded work (Narayanan & Moritz, 2015; Sterman, 1989a). As the value of β approaches 1, the decision rule shown in equations (2) and (3) begin to consider the *entire* inbound supply line with no or minimal discounting. Differing values of this parameter were used in previous studies to show how different levels of cognition in real human players of the Beer Game resulted in differing levels of inventory and ordering amplification. Correspondingly, the entities developed here, optimized to minimize system costs in the presence of human-like supply chain partners, act like the high cognition players seen in those previous studies (Narayanan & Moritz, 2015) by *completely* considering the inbound supply line when making ordering decisions.

## 4.3 Values of S' resembling a base-sock replenishment method:

The parameter S' maps approximately to the level of inventory on hand that the decision maker strives to maintain. Of interest, the values of S' arrived at by the optimized entities generally match a policy resembling a base-stock order method as expected in a full-information system with full rational entities (Clark & Scarf, 1960). The above optimization varies S' and α to create an effective base-stock level that minimizes the total cost of the system. For example, in the long-run steady state under the traditional customer order string of stepping from 4 to 8 units, and a balanced information and delivery day of 2 time periods each, we can expect a total of 16 units to be on-order in total (8 for each unit of time) and correspondingly 16 units in transit. Together, this represents 36 units that can be expected to flow into the on-hand inventory of the entity. Assuming incoming orders remain stable at 8 and outgoing orders match that number then maintaining a base-stock level of 36 is a realistic simplification of the full optimal policy.

## 4.4 Comparing Model-Based and Model-Free Approaches

Finally, examining Table 2 and Figure 6 help probe the boundary between model-based and model-free approaches and illustrates a surprisingly complementarity. Notably, the model-free DQN agent becomes more effect at reducing costs (relative to the model-based agent) under more noisy environments. Additionally, the model-free approach is more valuable at more highly decoupled, or central, positions in the supply chain (the wholesaler and the distributor here). Of note, these positions contributed approximately 55% of the real-world costs incurred by teams in runs of the Beer Game done for the incoming MIT Sloan MBA class in August of 2021, further implying that these results are valuable in a real-world setting. Despite the value of the DQN approach, the model-based approach is still surprisingly robust, and is generally cost reducing unless exposes to a highly noisy environment. Given the observations above about the interpretation of the parameters used in the model-based agents, this further lends support to prior work that emphasizes avoiding supply chain underweighting and trust amongst supply chain neighbors.

# 5   Limitations, Caveats and Future Work

The model-free approach above is still somewhat based on underlying assumptions of the environment in when choosing the architecture of the DQN, and thus could be labeled as a 'model-informed' model-free approach. This is not a critique, but rather a strength, of this approach and has allowed for the application of a dual DQN architecture in a novel context. However, a serious critique of the above work is that the model-based approach is not truly model-based in the strictest sense. The BFGS cost reducing routine treats the entire system as a function to be optimized, and thus is constrained by the model of the system that yields that functional form but does not truly capture a model of transition probabilities between states. To truly make the claim that this work probes the intersections and complementariness of model-based and model-free approaches, a true model-based approach, such as model predictive control, needs to be developed and incorporated into this research.

Development of a model predictive control scheme would not only provide a more concrete model-based example, but possibly would open more opportunities for empirically testing the agents developed herein. Both agents developed so far require pre-training on a system, after which they are static in their control policies. Model predictive control however should be able to be partially pretrained (by defining a 'default' control policy) but then updated online as ordering behaviors change. This would allow for this empirically grounded simulation work to perhaps be empirically verified by using each of these three agent types in a real run of the beer game.

Any empirical study has a confounding factor of trust of decision support systems of real human players and begs the question of if knowledge of the presence of an algorithmic intervention modifies human ordering behavior. This is beyond the scope of the current work and even the next steps of incorporating model predictive control schemes and the resultant empirical tests but could yield additional fruitful future work.